\documentclass[aps,showpacs,12pt,groupedaddress,%
superscriptaddress,amsmath,amssymb]{revtex4-1}

\usepackage[breaklinks = true,colorlinks = true,%
linkcolor = blue,citecolor = blue]{hyperref}
\usepackage[dvipdfmx]{graphicx}
\usepackage{xcolor}

\newcommand{\rme}{\mathrm{e}}
\newcommand{\rmi}{\mathrm{i}}
\newcommand{\rmd}{\mathrm{d}}
\newcommand{\re}{\mathrm{Re\,}}
\newcommand{\im}{\mathrm{Im\,}}

\begin{document}

%\preprint{XXXX}

\title{Featuring the topology with the unsupervised machine learning}

\author{Kenji Fukushima}
\email{fuku@nt.phys.s.u-tokyo.ac.jp}
\affiliation{Department of Physics, The University of Tokyo,
  7-3-1 Hongo, Bunkyo-ku, Tokyo 113-0033, Japan}
\affiliation{Institute for Physics of Intelligence (IPI), The University of Tokyo,
  7-3-1 Hongo, Bunkyo-ku, Tokyo 113-0033, Japan}

\author{Shotaro Shiba Funai}
\email{shotaro.funai@oist.jp}
\affiliation{Physics and Biology Unit, Okinawa Institute of Science and Technology (OIST), 
  1919-1 Tancha Onna-son, Kunigami-gun, Okinawa 904-0495, Japan}

\author{Hideaki Iida}
\email{iida@nt.phys.s.u-tokyo.ac.jp}
\affiliation{Department of Physics, The University of Tokyo,
  7-3-1 Hongo, Bunkyo-ku, Tokyo 113-0033, Japan}

\begin{abstract}
\vspace*{10pt}
  Images of line drawings are generally composed of primitive
  elements.  One of the most fundamental elements to characterize
  images is the topology; line segments belong to a category different
  from closed circles, and closed circles with different winding
  degrees are nonequivalent.  We investigate images with nontrivial winding
  using the unsupervised machine learning.  We build an autoencoder
  model with a combination of convolutional and fully connected neural
  networks.  We confirm that compressed data filtered from the trained
  model retain more than $90\%$ of correct information on the
  topology, evidencing that image clustering from the unsupervised
  learning features the topology.
\end{abstract}

\maketitle

%%%%%%%%%%%%%%%%%%%%%%%%%%%%%%%%%%%%%%%%%%%%%%%%%%
\section{Introduction}
%%%%%%%%%%%%%%%%%%%%%%%%%%%%%%%%%%%%%%%%%%%%%%%%%%

Brains ingeniously function with networks of neurons.  For
understanding of intrinsic brain dynamics, physicists would favorably
decompose such an integral system into irreducible elements,
so that we can analyze relatively simpler function of each building
element that takes rather primitive actions.  Then, numerical simulations
on computer are handy devices to test if a postulated mechanism of
brains should go as expected.  Such modeling embodies non-equilibrium
processes of brains, which is an approach acknowledged commonly as
computational neuroscience.  Besides, a hypothesis called quantum
brain dynamics implements quantum fluctuations and Nambu-Goldstone
bosons for brain sciences~\cite{ricciardi1967brain,jibu1995quantum}
(for discussions for/against quantum phenomena in brain dynamics, see
Ref.~\cite{Jedlicka2017}), which bridges a devide between
computational neuroscience and modern physics. 

In contrast to such ``off-equilibrium'' problems, in the language of
physics, perception and recognition are ``static'' problems.  For the
latter problems, model-independent research tools are available for
computer simulations.  That is, the machine learning enables us to
emulate the neural structure of brains on computer.
One of intriguing attributes of the machine learning, particularly with
deep neural networks (NNs), is that any nonlinear mapping can be
represented by data transmission through multiple hidden
layers~\cite{lecun2015deep}.

These days we have witnessed tremendous progresses in the field of
image recognition and classification by means of the machine learning.
In particular the progress has been driven by Convolutional Neural
Network (CNN)~\cite{lecun1989backpropagation}, which was originally
proposed as a multi-layer neural network imitating animal's visual
cortex~\cite{fukushima1980neocognitron}.  The CNN has become the most
common approach for high-level image recognition since an overwhelming
victory of AlexNet~\cite{krizhevsky2012imagenet} at ``ImageNet Large
Scale Visual Recognition Challenge 2012.''
Challenges of minimizing inter-class variability, reducing error rate,
achieving large-scale image recognition, etc are ongoing improvements
and they are all crucial steps for practical usages.

Physicswise, the image handling with the deep learning has proved its
strength in identifying phase transitions.  Some successful attempts
are found in Ref.~\cite{doi:10.7566/JPSJ.85.123706} for
two-dimensional systems analyzed by the supervised learning,
Ref.~\cite{doi:10.7566/JPSJ.86.044708} for its extension to
three-dimensional systems, and Ref.~\cite{Funai:2018esm} for
statistical systems studied by the unsupervised learning.   Here, we
point out an essential difference between the supervised and
unsupervised learning;  the former is useful for regression and
grouping problems, while the latter efficiently makes feature
extraction and clustering of data.  Interestingly, similarity between
the unsupervised learning and the renormalization group in physics has
also been investigated, see Ref.~\cite{Iso:2018yqu}.

In the context of image recognition, at the same time, a distinct
direction toward more fundamental research would be as important to
demystify blackboxed artificial intelligence, which may be somehow
beneficial for so-called explainable artificial
intelligence~\cite{DBLP,8400040}.  The fundamental question of our
interest in the present work is what would be the simplest element of
images that categorizes those images into representative clusters.
For the sake of image clustering, a useful mathematical notion, which
underlies modern physics, has been developed known as the ``topology''
theorized into the form of homotopy.  The most well-known example is
that a mug with one handle and a torus-shaped donut belong to the same
grouping class; the shape can be smoothly deformed from one to the
other, and they are of the same homotopy type.  In this work we report
leading-edge results from our simulations with the CNN supporting an
idea that the topology is critical information for image feature
extraction and clustering.

%%%%%%%%%%%%%%%%%%%%%%%%%%%%%%%%%%%%%%%%%%%%%%%%%%
\section{Topology and the winding number}
%%%%%%%%%%%%%%%%%%%%%%%%%%%%%%%%%%%%%%%%%%%%%%%%%%

\begin{figure}
  \centering 
  \includegraphics[width=0.6\textwidth]{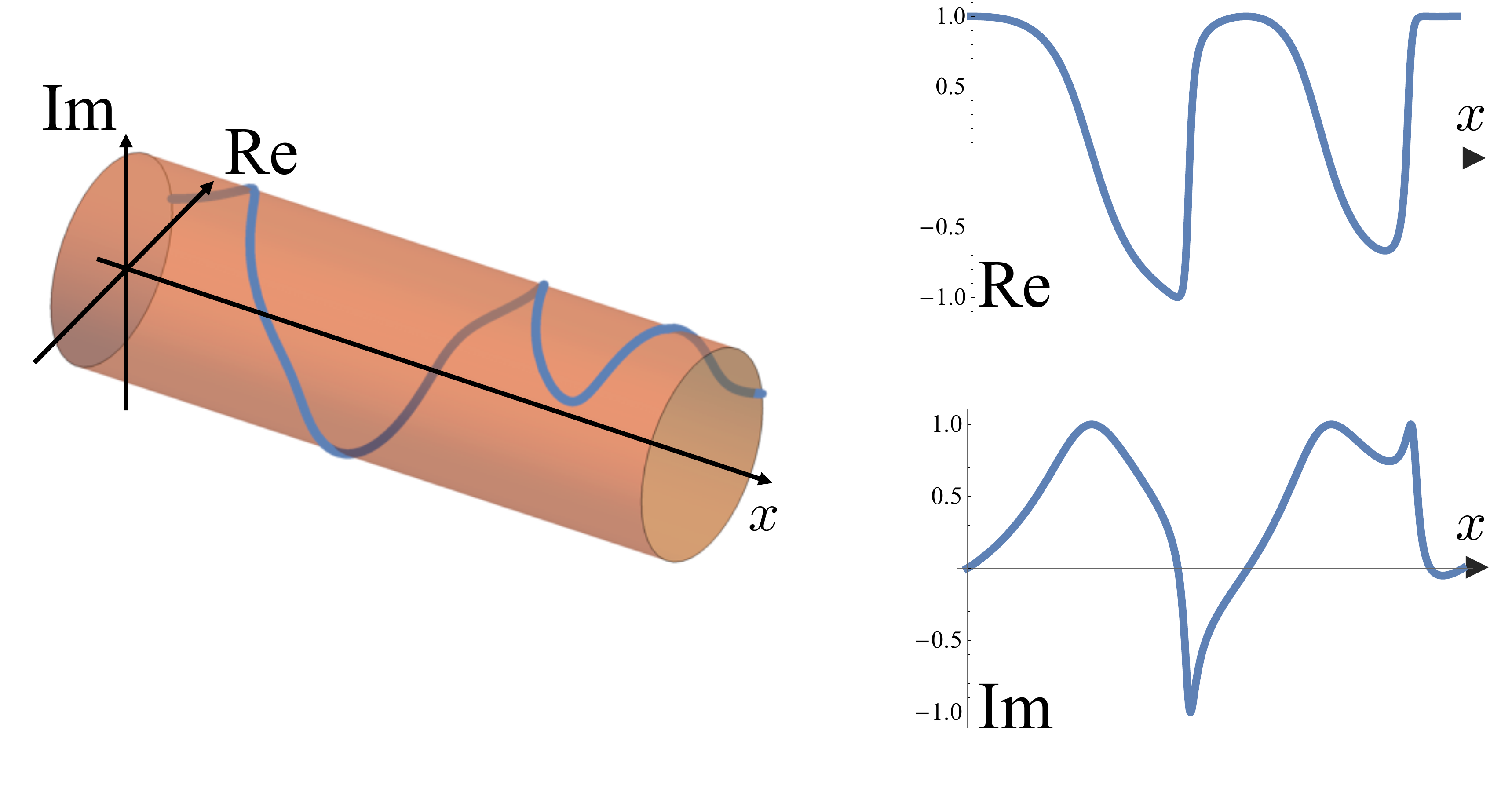}
  \caption{Schematic illustration of $\pi_1(S^1)$ realized by
    $\phi(x)=\rme^{\rmi\theta(x)}$.  In the left representation the
    behavior of $\theta(x)$ is manifest and $n_W=1$ is easily
    concluded, but in our simulation, only $(\re\phi,\im\phi)$ as
    shown in the right is given to the NN model.}
  \label{fig:schematic}
\end{figure}

The topology is classified by the homotopy group in mathematics.  The
simplest example is what is called the fundamental homotopy group
denoted as $\pi_1(S^1)=\mathbb{Z}$ associated with a mapping from
$S^1$ (i.e., one dimensional unit sphere) to another $S^1$ and an
integer $n_W \in\mathbb{Z}$ corresponds to the winding number.  To
demonstrate the idea concretely, let us consider the following
function on $S^1$ of U(1),
\begin{equation}
  \phi(x) = \rme^{\rmi \theta(x)}
  = \cos\theta(x) + \rmi \sin\theta(x)\,. 
\end{equation}
If $x$ is a coordinate on a circle with period $L$, the above function
represents a mapping
from $S^1$ in coordinate space to $S^1$ on Gauss' plane with Euler's
angle $\theta$ (which is also called the ``lift'' in homotopy theory).
While $x$ travels around from 0 to $L$ under a condition,
$\phi(0)=\phi(L)$, Euler's angle $\theta$ should return to the
original position modulo $2\pi$.  The winding number associated with
the above function (or the ``degree'' of this function) reads,
\begin{equation}
  n_W = \frac{\theta(L)-\theta(0)}{2\pi}
  = \frac{\ln[\phi(L) / \phi(0)]}{2\pi\rmi}
  = \frac{1}{2\pi\rmi} \int_0^L \rmd x\, \phi^{-1}(x) 
  \frac{\rmd\phi(x)}{\rmd x} \,.
  \label{eq:winding}
\end{equation}
Figure~\ref{fig:schematic} schematically illustrates one winding
configuration of $\phi(x)$ having $n_W=1$.

% --- figure ---%
% --- figure ---%

For us, human-beings, it would be an elementary-class exercise to
discover a counting rule of $n_W$.  If we were given many different
images with correct answers of $n_W$, it would be just a matter of
time for us to eventually find a right counting rule out.  This
description is nothing but the machinery of the
\textit{supervised learning}. 
Interestingly, it has been reported that the deep neural network
trained by the supervised learning correctly reproduced $n_W$ of
$\pi_1(S^1)$~\cite{zhang2018machine}.  There, the machine discovered
a nice fit of the formula~\eqref{eq:winding} only from the information
of given $(\re\phi,\im\phi)$ and $n_W$, but not referring to
$\theta(x)$ directly.  In the present work, we are taking one step
forward;  we would like to see the NN model not only optimizing a fit
from the supervised learning but featuring the topology from the
\textit{unsupervised learning}.

More specifically, we would like to think of classification of many
images without giving the answers of $n_W$.  It would be very
intriguing to ask such a question of how the CNN makes clustering of
differently winding data, which would be a prototype model of how our
brains categorize images based on the topological characterization.
Thus, the unsupervised learning as adopted in this work should tell us
surpassing information than the supervised learning for the purpose to
dissect the topological contents.

%%%%%%%%%%%%%%%%%%%%%%%%%%%%%%%%%%%%%%%%%%%%%%%%%%
\section{Results and Discussions}
%%%%%%%%%%%%%%%%%%%%%%%%%%%%%%%%%%%%%%%%%%%%%%%%%%

In the numerical procedure we represent $\phi(x)$ by a sequence of
numbers on discretized $x$ with $L=128$ grids,
i.e., we generate $2\times 128=256$ sized data of
$(\re\phi_i,\im\phi_i)$ with $i=1,\dots,L$ under the periodic
boundary condition;
$(\re\phi_L,\im\phi_L)=(\re\phi_1,\im\phi_1)$.
These 256 numbers consist of the input data onto the CNN side.

% --- figure ---%
\begin{figure}
  \centering 
  \includegraphics[width=0.5\textwidth]{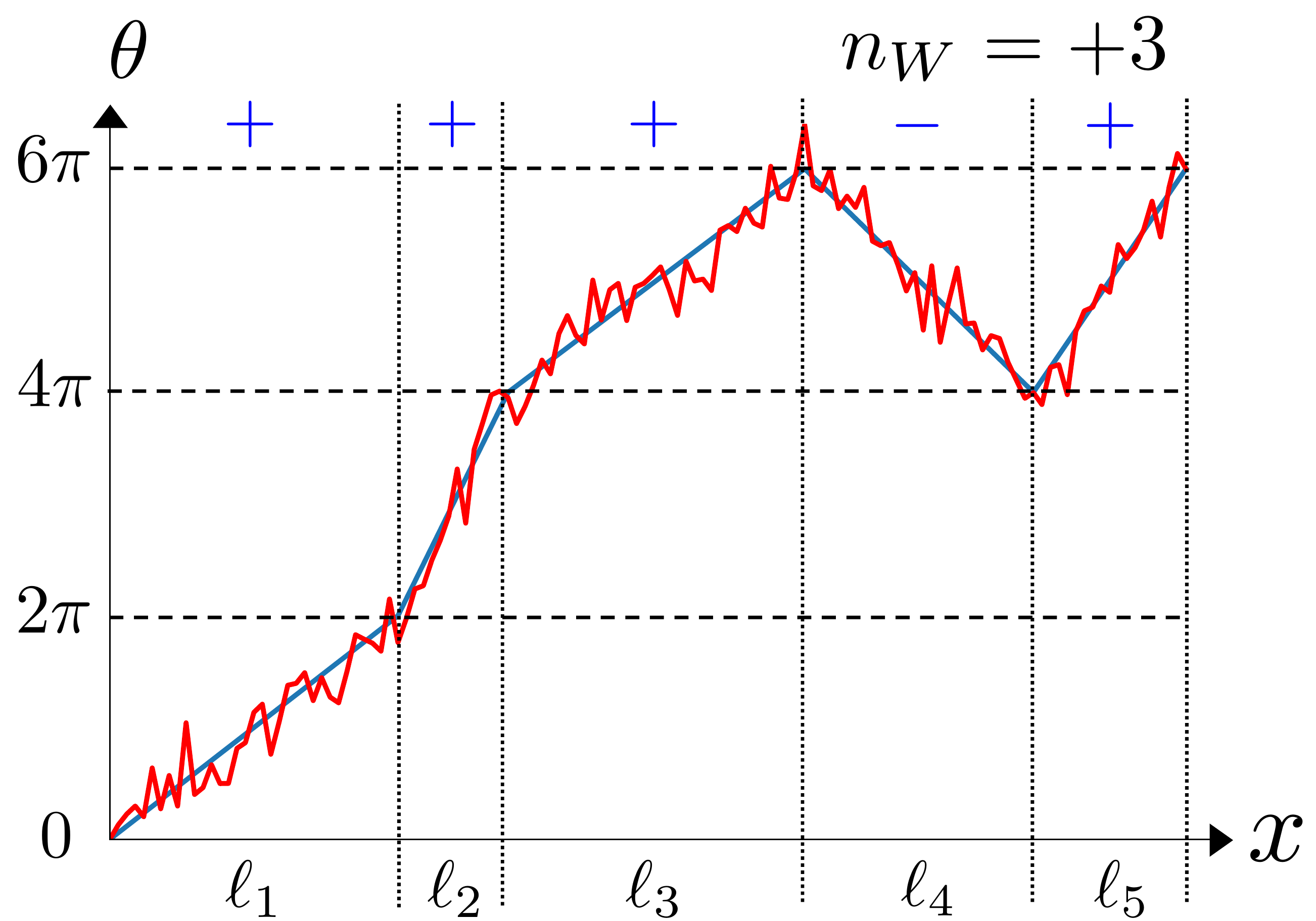}
  \caption{One example of generated data for $(+,+,+,-,+)$ with
    $n_W=+3$.}
  \label{fig:data}
\end{figure}
% --- figure ---%

We prepare the training and test data randomly.  We will give more
detailed explanations on the numerical procedure in Method.  Each data
consists of distinct $N_s$ segments along $x$ with either positive
or negative winding, where the segment lengths, $\ell_m$, are chosen
randomly, where $\displaystyle\sum_{m=1}^{N_s} \ell_m=L$ should be kept.  For the
data used in this work, we take account of $N_s=0,1,2,3,4$, and $5$.
We then assign positive and negative winding randomly to each segment,
which is symbolically labeled as $(p_1,p_2,\dots,p_{N_s})$ with
$p_m=\pm 1$, where $+$ (and $-$) stands for positive (and negative,
respectively) winding.  For example, $(+,+,+,-,+)$ for $N_s=5$ is a
configuration with net winding, $\displaystyle n_W=\sum_{m=1}^{N_s} p_m=+3$.  In
$m$-th segment we first postulate that $\theta(x)$ linearly changes its
value by $2\pi p_m$.  Then, zigzag lines are distorted with random
noises to enhance the learning quality and thus the adaptivity.
Figure~\ref{fig:data} depicts an example of generated data with a
choice of $(+,+,+,-,+)$.  With $N_s$ up to 5
$2^0+2^1+2^2+2^3+2^4+2^5=63$ winding patterns are possible, and $n_W$
can take a value from $-5$ to $+5$.  This setup is for the moment sufficiently
general for our goal to check performance of the topology detection.

The unsupervised learning utilizes the autoencoder~\cite{Hinton504};  
it first encodes the data compressed into smaller number of neurons in
the CNN (in our case, 16 sites\,$\times$\,4 filters from original 256 sites)
and then decodes the compressed data with fully connected NN into the
original size.  We repeat such encoding and decoding processes to
minimize the loss function measured by the squared difference between
the original input data and the coarse-grained output data.  The
learning process optimizes the filters in the CNN encoder and
simultaneously the weights in the NN decoder.

\begin{figure}
  \centering 
  \includegraphics[width=0.42\textwidth]{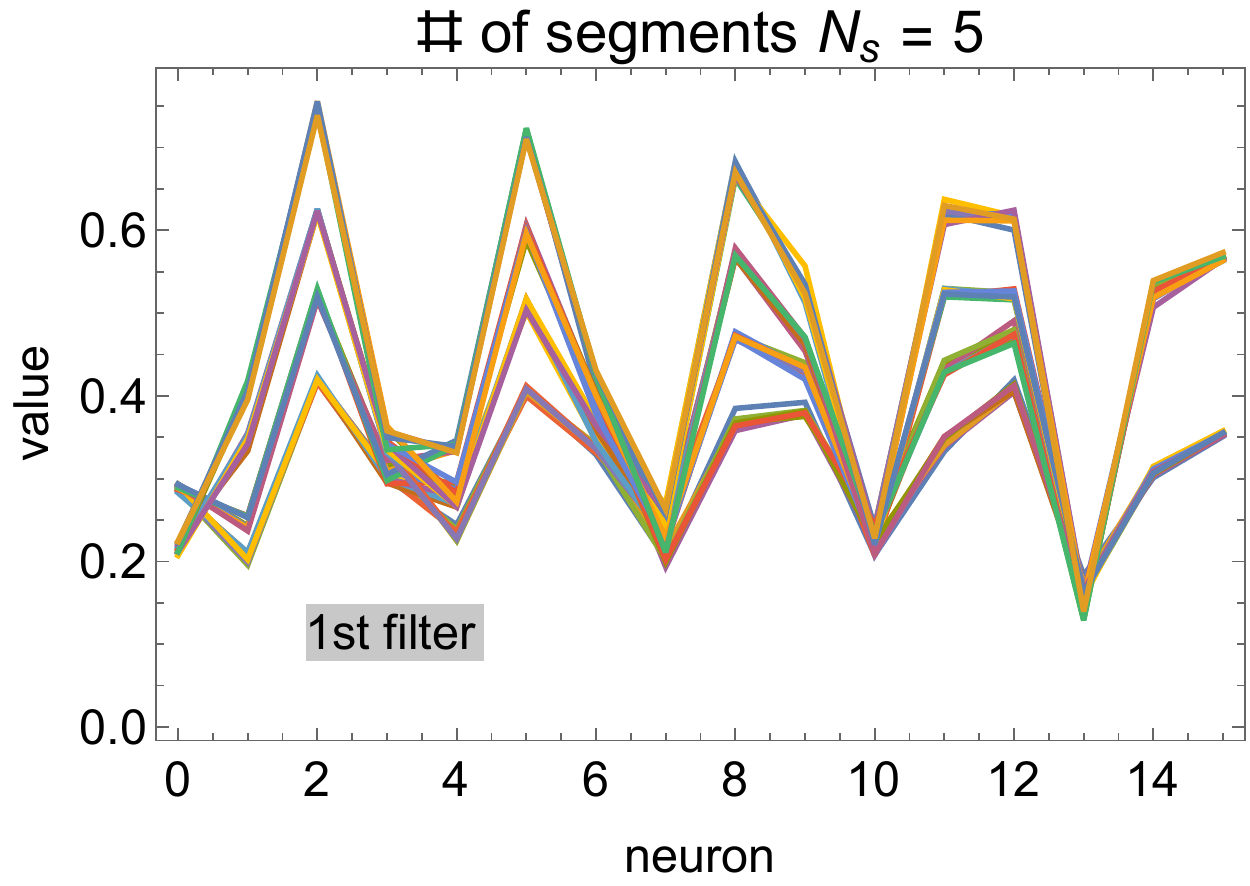} \hspace{0.3em}
  \includegraphics[width=0.52\textwidth]{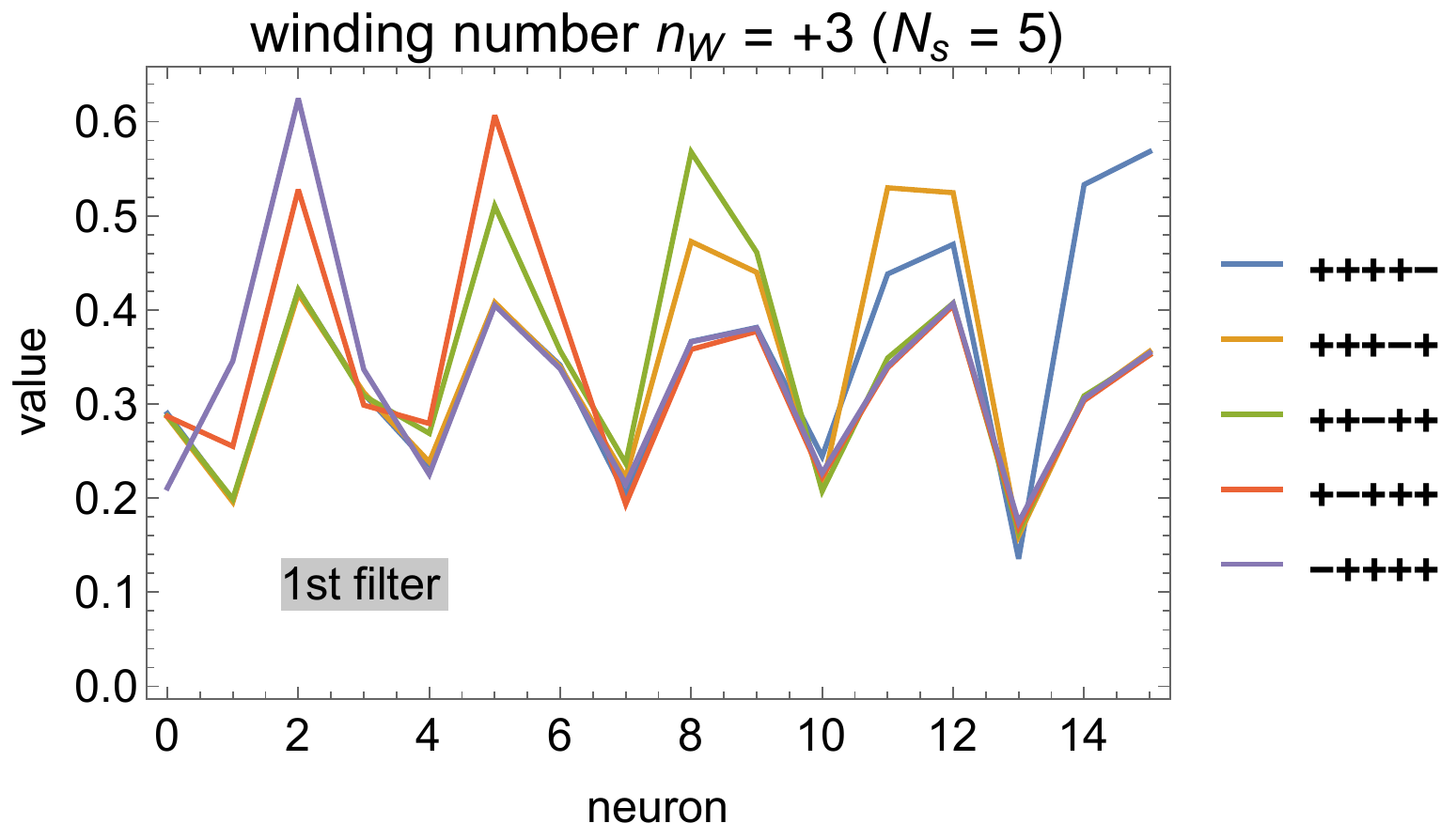}
  \caption{Examples of feature maps on the deepest CNN layer.  (Left)
    $N_s=5$ and all possible (i.e., $2^5=32$) winding patterns
    averaged over 1,000 test data.  (Right) $N_s=5$ with $n_W=+3$ fixed and
    five winding patterns averaged over 1,000 test data.}
  \label{fig:pool}
\end{figure}

We shall see the results from our NN model that has been
optimized by the unsupervised learning with the training dataset of 63
winding patterns times 1,000 randomly generated data (i.e., 63,000
data in total).  We input the test data into the optimized NN model
and observe feature maps of 16 neurons convoluted with 4 filters on
the deepest CNN layer.  Figure~\ref{fig:pool} summarizes sampled
feature maps from the 1st filter.  The left plot is for $N_s=5$ with
all possible winding patterns averaged over 1,000 test data.  The
right plot particularly picks up the averaged feature maps for five
different windings with $N_s=5$ and $n_W=+3$.  At a glance one may
think that the behavior looks like coarse-grained $\re\phi(x)$ or
$\im\phi(x)$ with segment lengths normalized.

% --- figure ---%
% --- figure ---%

Now, the most fascinated question is whether the compressed data on 16
sites convoluted with 4 filters could retain information on the
topology or not, and if yes, how we can retrieve it.  From the left of
Fig.~\ref{fig:pool} it is obvious that the peak heights reflect
different winding patterns.  In fact, the averaged feature maps
exhibit a clear hierarchy of four separated heights with one-to-one
correspondance to the winding sequence;  that is, the peak heights
increase with sequential windings as
\begin{equation}
  (+,+) ~~<~~ (+,-) ~~<~~ (-,+) ~~<~~ (-,-)\,,
  \label{eq:hierarchy}
\end{equation}
and the height at the far right end is determined solely by $\pm$, which
is due to zero-padding in the convolution to keep the data size.

For example, if we see the far left peak around the 2nd neuron in the
right of Fig.~\ref{fig:pool}, $(-,+,+,+,+)$ has the highest peak,
$(+,-,+,+,+)$ the second highest, and others are degenerated in accord
with Eq.~\eqref{eq:hierarchy}.  Also, we notice in the right of
Fig.~\ref{fig:pool} that, for $(+,+,+,+,-)$, three consecutive short
peaks appear from the left, one middle peak follows, and then one tall
peak sits at the right end.  The short peaks correspond to $(+,+)$ of
$((+,+),+,+,-)$, $(+,(+,+),+,-)$, and $(+,+,(+,+),-)$.  The middle
peak corresponds to $(+,+,+,(+,-))$ and the tall peak is sensitive
only to $-$ of $(+,+,+,+,-)$.

\begin{figure}
  \centering 
  \includegraphics[width=0.45\textwidth]{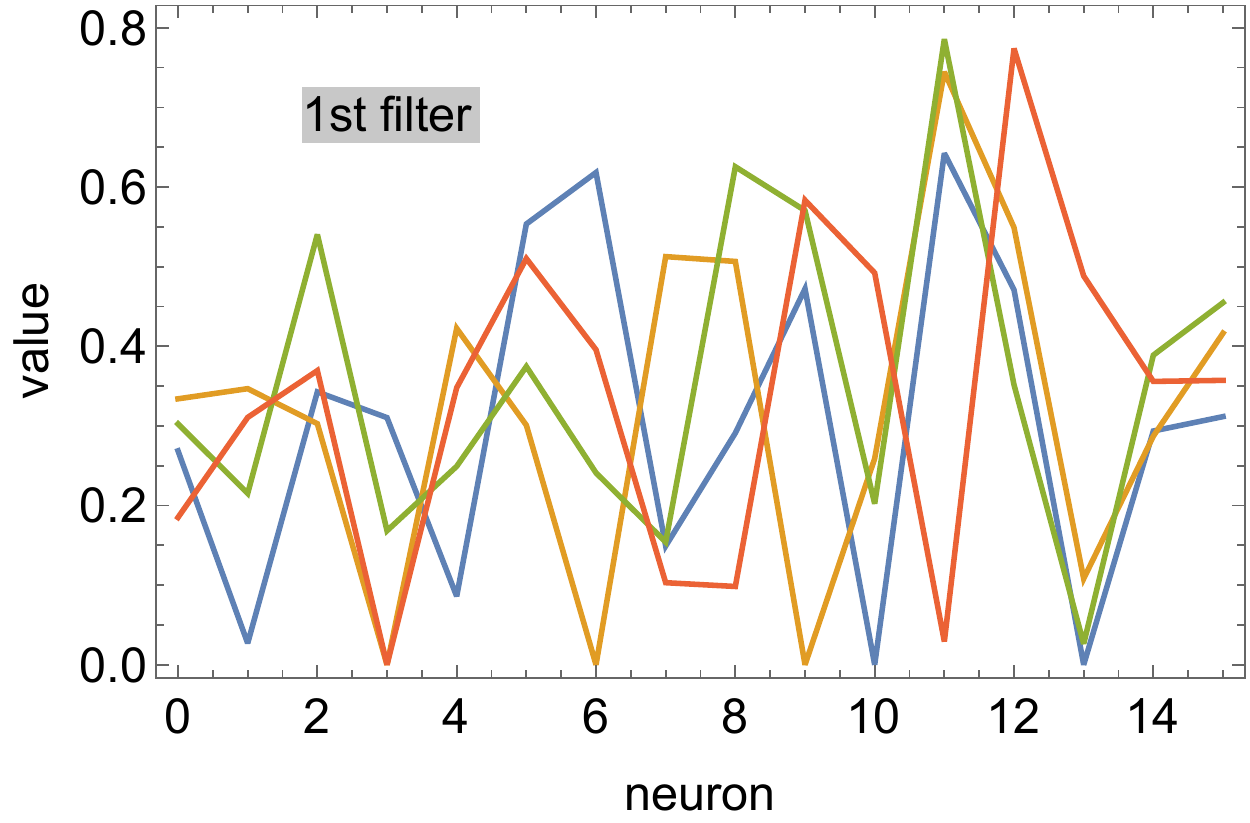} \hspace{1em}
  \includegraphics[width=0.45\textwidth]{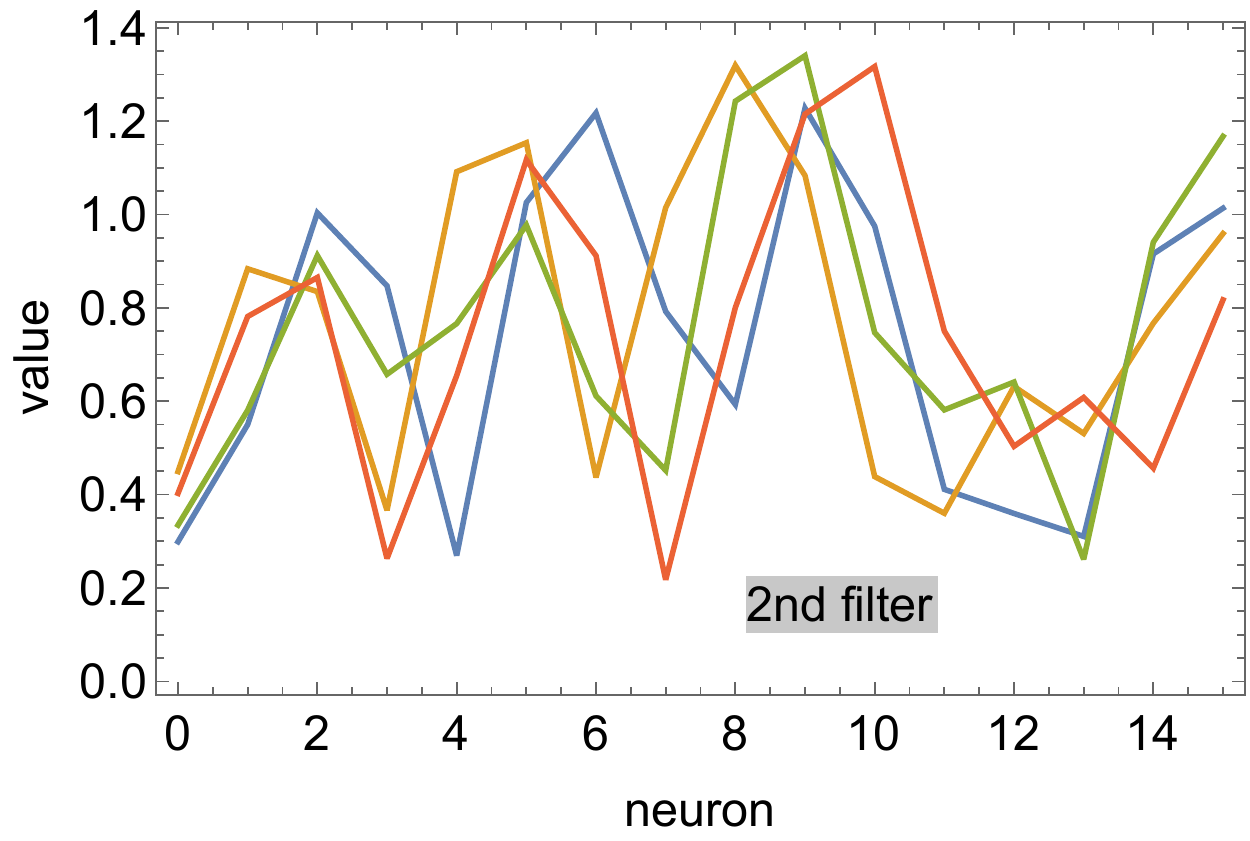} \\
  \includegraphics[width=0.45\textwidth]{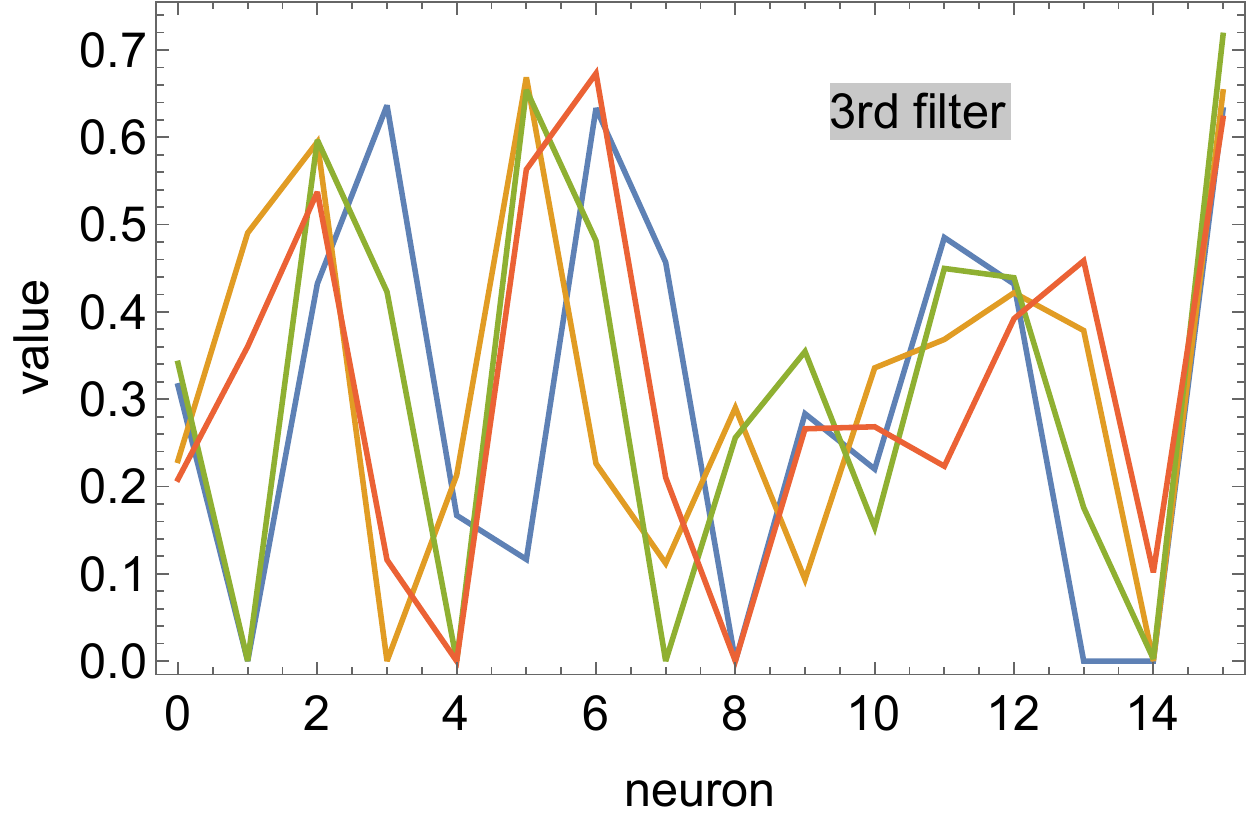} \hspace{1em}
  \includegraphics[width=0.45\textwidth]{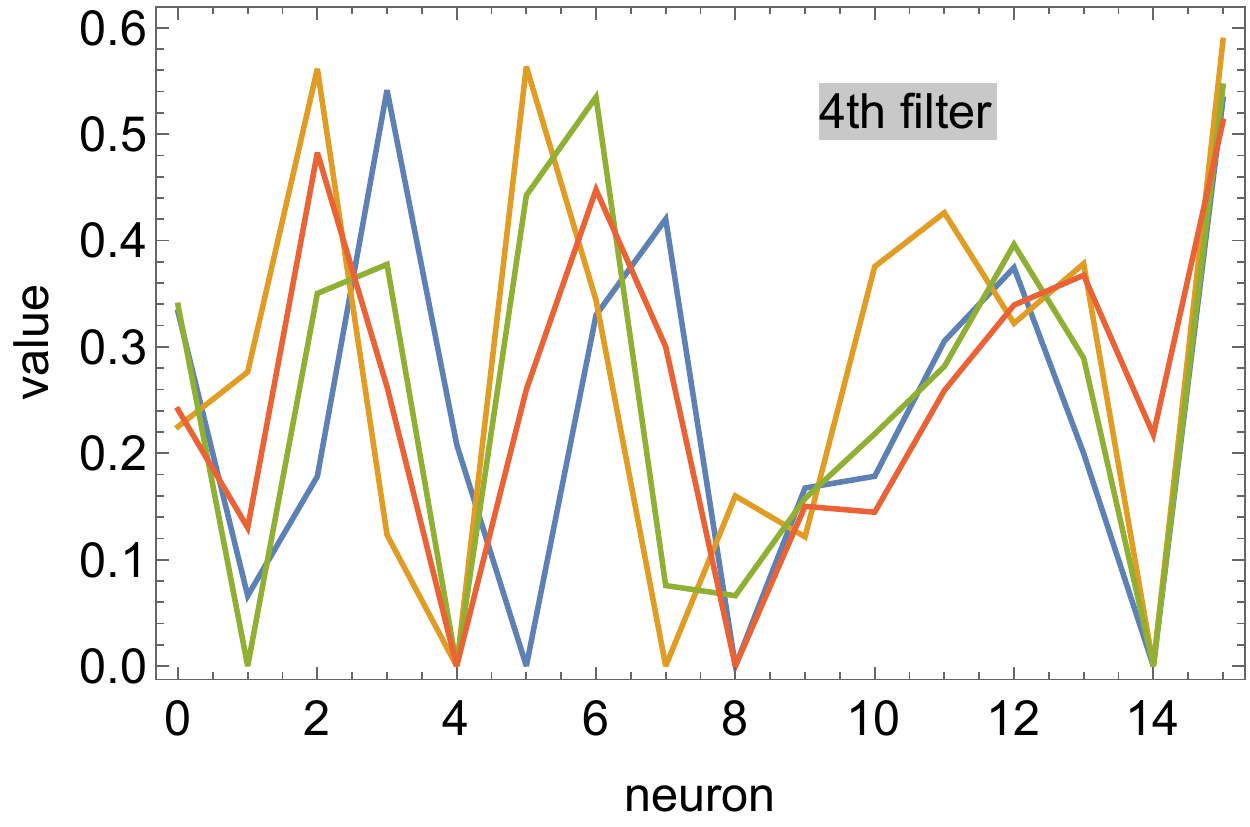}
  \caption{Feature maps from the 1st filter (top-left), the 2nd filter
    (top-right), the 3rd filter (bottom-left), and the 4th filter
    (bottom-right) for $(+,+,+,-,+)$ data without taking the average.
  Four randomly selected data are shown by four different colors.}
  \label{fig:notaved}
\end{figure}

One might have thought that such a clear hierarchy as in
Fig.~\ref{fig:pool} is visible only after taking the average.
This is indeed the case, as exemplified in Fig.~\ref{fig:notaved};  the
feature maps for one test data do not always show prominent peaks with
well separated heights.  Nevertheless, surprisingly, we can see that
such fluctuating coarse-grained data in Fig.~\ref{fig:notaved} still
retain information on the topology!

% --- figure ---%
% --- figure ---%

\begin{figure}
  \centering 
  \includegraphics[width=0.7\textwidth]{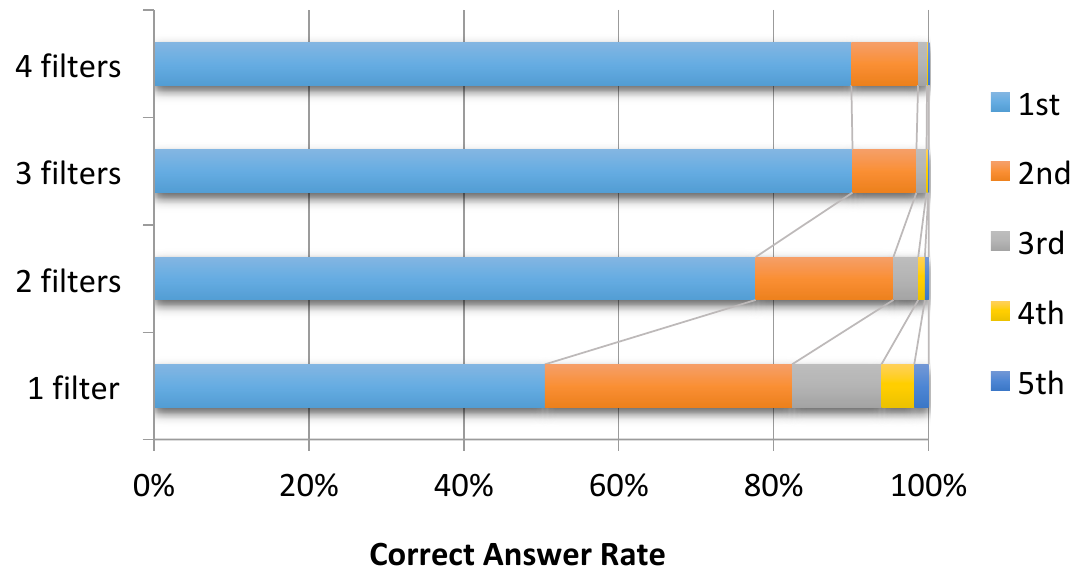}
  \caption{Correct answer rate for $n_W$ guessed from the feature
    maps.  For each of 63,000 test data we have checked where the correct
    winding number ranks (1st, 2nd, ... in the legend) in the
    probability output from the supervised NN model. }
  \label{fig:score}
\end{figure}

It is impossible for our eyes to recognize any topological contents
from Fig.~\ref{fig:notaved}, so we will ask for a help of another
machine learning device.  For each of 63,000 training data, we have
such feature maps like Fig.~\ref{fig:notaved} and also the
corresponding $n_W$.  We can then perform the supervised learning to
train a fully connected NN (with one hidden layer) such that the
output gives the probability distribution of guessed $n_W$ (out of
$-5,\ldots,5$) in response to the input of feature maps.
Figure~\ref{fig:score} is the correct answer rate.  If we input the
feature maps from only 1 filter, the most-likely $n_W$ hits the
correct value at the rate of $50\%$, and the second-likely one at the
rate of $31\%$.  If we use the feature maps from 2 filters, the
available information is doubled, and the correct answer rate of the
most-likely $n_W$ becomes $78\%$.  Amazingly, for the feature maps
from 3 filters, it increases up to $90\%$!  There is almost no
difference between the 3-filter and the 4-filter results, and it seems
that the correct answer rate is saturated at $90\%$.

%\funai{
%As another way to get the topology information our autoencoder retains,
%we can calculate the winding number of the coarse-grained output data 
%using Eq.\,(\ref{eq:winding}), and compare it with the true winding number.
%The correct answer rate is $\sim 83\%$, when we input all the 4 filters,
%adjust the absolute values of output data to 1 with their arguments kept,
%and round the calculated winding numbers to integers.
%This result is slightly lower than Fig.~\ref{fig:score}, which should be
%because the integration in Eq.\,(\ref{eq:winding}) is discretized.}

To summarize, from these results and analyses, we can conclude that
the coarse-grained data of the feature maps through the unsupervised
learning do retain information on the topology.  In other words, the
unsupervised learning with autoencoder can provide us with a
nontrivial machinery to compress data without losing the topological
contents.

% --- figure ---%
% --- figure ---%

%%%%%%%%%%%%%%%%%%%%%%%%%%%%%%%%%%%%%%%%%%%%%%%%%%
\section{Discussion}
%%%%%%%%%%%%%%%%%%%%%%%%%%%%%%%%%%%%%%%%%%%%%%%%%%

We demonstrated that the unsupervised machine learning of images makes
feature extraction without losing information on the topology.  This
evidences that the winding number corresponding to the fundamental
group, $\pi_1(S^1)$, should be one of the essential indices that
characterize image clustering.  We trained an autoencoder with the CNN
and the fully connected NN for the unsupervised learning using
randomly generated data of functions from $x$ on $S^1$ to a
U(1) element.  We found that the averaged feature maps on the deepest
CNN layer show a clear hierarchy pattern of the configurations with
one-to-one correspondence to the winding  sequences.  With help of the
supervised learning technique we also revealed that feature maps for
each image data look coarse-grained images and such compressed data
retain information on the topology correctly.

The extension of the present work, i.e., the unsupervised learning of
higher-dimensional images with nontrivial winding would be quite
interesting.  We note that the supervised machine learning has been
utilized for $\pi_2(S^2)$~\cite{zhang2018machine}, but as we
illustrated in this work, the unsupervised machine learning would be
more interesting.  Implications from this extension include intriguing
applications in quantum field theories.  In fact, some classical
solutions of the equations of motion in quantum field theory are
topologically stabilized.  In such cases the field configurations are
classified according to the winding number.  For
representative examples, $\pi_2({\rm SU(2)/U(1)})=\mathbb{Z}$ for
monopoles, $\pi_3({\rm SU(2)})=\mathbb{Z}$ for Skyrmions (with no
time-dependence), and $\pi_3({\rm SU(2)})=\mathbb{Z}$ for instantons
(with the fields at infinity identified) in pure Yang-Mills
theory~\cite{belavin1975pseudoparticle,hooft1994computation}.
Actually, it is a long-standing problem how to visualize the
topological contents in quantum field theories.  The numerical lattice
simulation is a powerful tool to solve quantum field theories, and
field configurations should in principle contain information on the
topology.  Some algorithms to extract topologically nontrivial
configurations such as monopoles, Skyrmions, and instantons have been
proposed~\cite{berg1981dislocations,teper1986topological,
ilgenfritz1986first}.  The most well-known approach, i.e., the cooling
method has a serious flaw, however.  If the cooling is applied too
many times, the topology is lost and the field configuration would become
trivially flat.  Therefore, the cooling procedures should be stopped at
some point, and this artificial termination of the procedures causes
uncertainties.  Alternatively, we would emphasize that the compression
of field configuration images by means of the unsupervised machine
learning is a promising candidate for the superior smearing algorithm
not to lose the topological contents.  We are testing this idea in
simple two-dimensional lattice model, namely, $\mathbb{C}P^{N-1}$
model that has $\pi_2(S^2)$ instantons.

Mathematically, it would be also a very interesting question to
consider not only the homotopy groups but also the homology $H_n(X)$
of a topological space $X$ defined by a coset of cycles in $n$
dimensions over boundaries of $(n+1)$-dimensional elements.  For
example, $H_0(X)$ counts the number of connected drawings of images,
and $H_1(X)$ counts the number of loops of images, etc.  This
direction of research is now ongoing.

\begin{acknowledgments}
We thank Yuya~Abe and Yuki~Fujimoto for
discussions.  K.~F.\ was supported by Japan Society for the Promotion
of Science (JSPS) KAKENHI Grant No.\,18H01211.
\end{acknowledgments}

%%%%%%%%%%%%%%%%%%%%%%%%%%%%%%%%%%%%%%%%%%%%%%%%%%
\appendix
\section*{Method}
%%%%%%%%%%%%%%%%%%%%%%%%%%%%%%%%%%%%%%%%%%%%%%%%%%

\subsection*{Input data of winding patterns}

Input data $\phi_i = e^{\rmi\theta_i}$ ($i=1,\ldots,L$), 
including training data and test data,
are generated in the following way.
Note that since the data $\phi_i$ are complex numbers,
we actually input the combinations of their real and imaginary parts
 ($\re \phi_i, \im \phi_i$).

First we impose the periodic boundary condition:
\begin{equation}
\re\phi_1 = \re\phi_L,\quad
\im\phi_1 = \im\phi_L.
\end{equation}
Then we divide $L$ sites into $N_s$ segments,
and length of each segment $\ell_m$ ($m=1,\ldots,N_s$)
is randomly chosen as 
\begin{equation}
\ell_m = \frac{L-1}{N_s}\left[1+0.4(\xi_m-\xi_{m-1})\right]
\end{equation}
where $\xi_m$ is a random number from a uniform distribution in the open interval $(-1,1)$.
This means the lengthes in all the segments satisfy 
$0.2 \frac{L-1}{N_s} < \ell_m < 1.8 \frac{L-1}{N_s}$.
We set $\xi_0 = \xi_{N_s} = 0$ so that the total length $L$ is kept.
To be exact, the length $\ell_m$ should be an integer, 
so the right hand side is rounded to an integer.

In each segment $m$,
the angle $\theta_i$ is composed of a linear part from 0 to $\pm 2\pi$ 
and a random noise:
\begin{equation}
\frac{\theta_i}{2\pi} = p_m \frac{i-i_0}{\ell_m} + 0.1 \zeta_i
\end{equation}
where 
$p_m=\pm 1$ is the winding direction and $i_0 = 1+\sum_{m'=1}^{m-1} \ell_{m'}$.
The random noise $\zeta_i$ is from a Gaussian distribution with mean 0 and variance 1.

In our experiment, we set $L=128$ and $N_s=0,1,\ldots,5$.
Then all the combinations of winding directions $p_m$ have 
$\sum_{N_s=0}^5 2^{N_s} = 63$ patterns.
For each winding pattern, we generate 1,000\,$+$\,1,000 input data 
with the parameters $(\xi_m, \zeta_i)$ chosen randomly.
The first 1,000 data (in total 63,000 data) is the training data, 
which is used for training our autoencoder and supervised NN.
The other 1,000 data is the test data for analyzing the feature maps
in the autoencoder (see Figs.~\ref{fig:pool} and \ref{fig:notaved}) and
the output from the supervised NN (see Fig.~\ref{fig:score}).

\subsection*{Autoencoder}

% --- figure ---%
\begin{figure}
  \centering
  \includegraphics[width=0.9\textwidth]{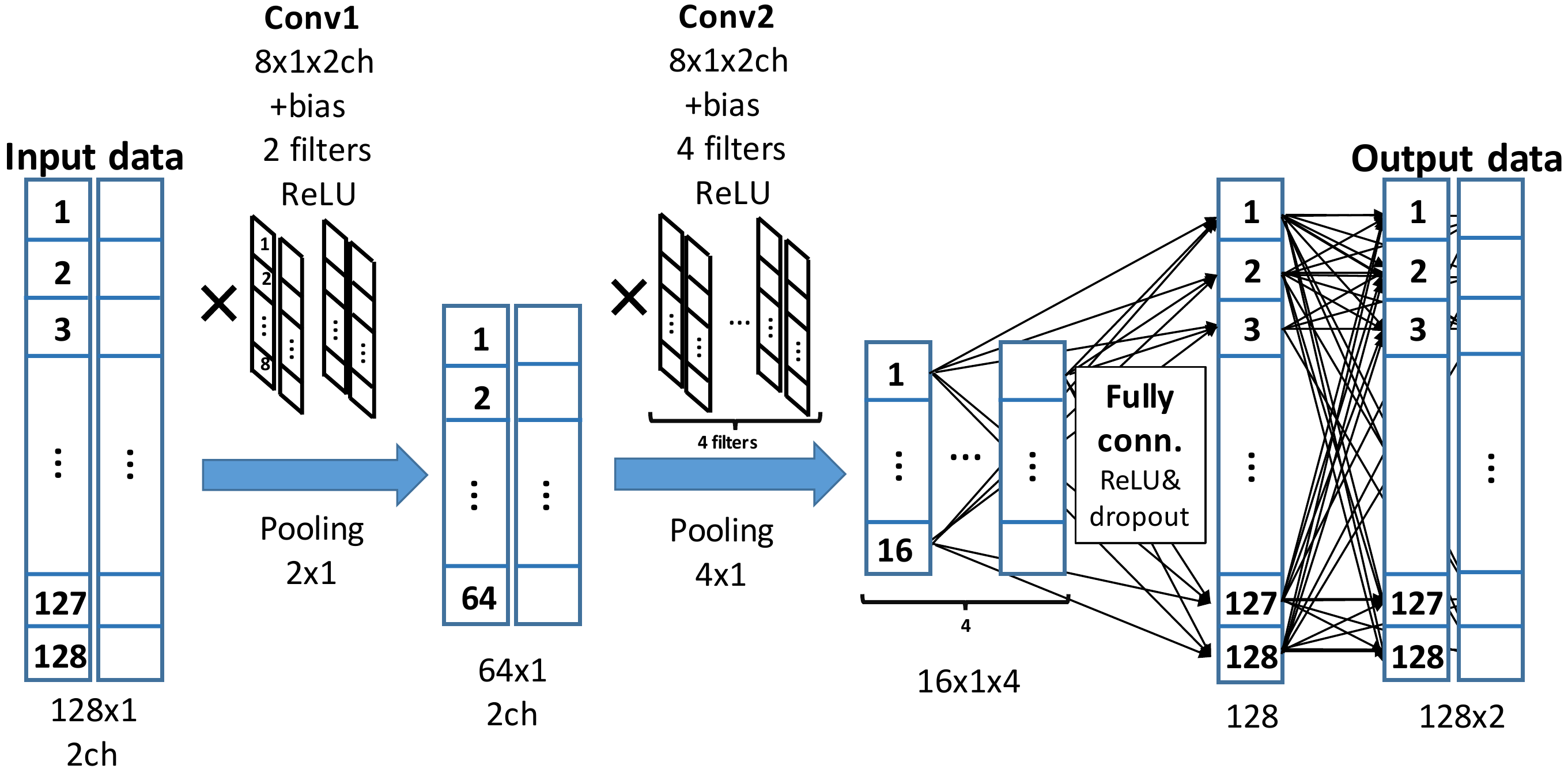}
  \caption{Schematic figure {of} % to picturize 
               the autoencoder used for our unsupervised learning.}
  \label{fig:cnn}
\end{figure}
% --- figure ---%

The autoencoder for our unsupervised learning consists of the CNN (encoder) and 
the fully-connected NN (decoder).
A schematic figure of our autoencoder is shown in Fig.~\ref{fig:cnn}.
We made a training code using TensorFlow~\cite{199317}.

The CNN encoder has two layers. 
In the both layers, we use the convolution with $8\times 1 (\times 2$ channel) 
sized filters and stride 1.
We also use the zero padding to keep the size of data, 
and the ReLU as an activation function.
In the convolution part, 
difference between the first and second layers is only the number of filters.
After the convolution, our encoder has the pooling part in each layer.
We use the max pooling with $2\times 1$ sized window and stride 2 
for each channel in the first layer.
This pooling compresses the data size into $1/$stride of the original size.
The second layer has $4\times 1$ sized window and stride 4, then as a result,
the CNN encoder outputs the feature map with $L/2/4=16$ sites 
(for each filter out of 4 filters),
as we saw in Fig.~\ref{fig:pool}.

The fully-connected NN decoder has two layers, too.
In the first layer, % ($16\times 4 \to 128$),
we use the dropout method with probability 0.5 to avoid overlearning,
then use the ReLU again.
The final layer has no dropout and no activation function, then outputs
coarse-grained data $(\varphi_{i,1}, \varphi_{i,2})$ with the same size 
as the input data $(\re\phi_i, \im\phi_i)$.

\begin{figure}
  \centering
  \includegraphics[width=0.7\textwidth]{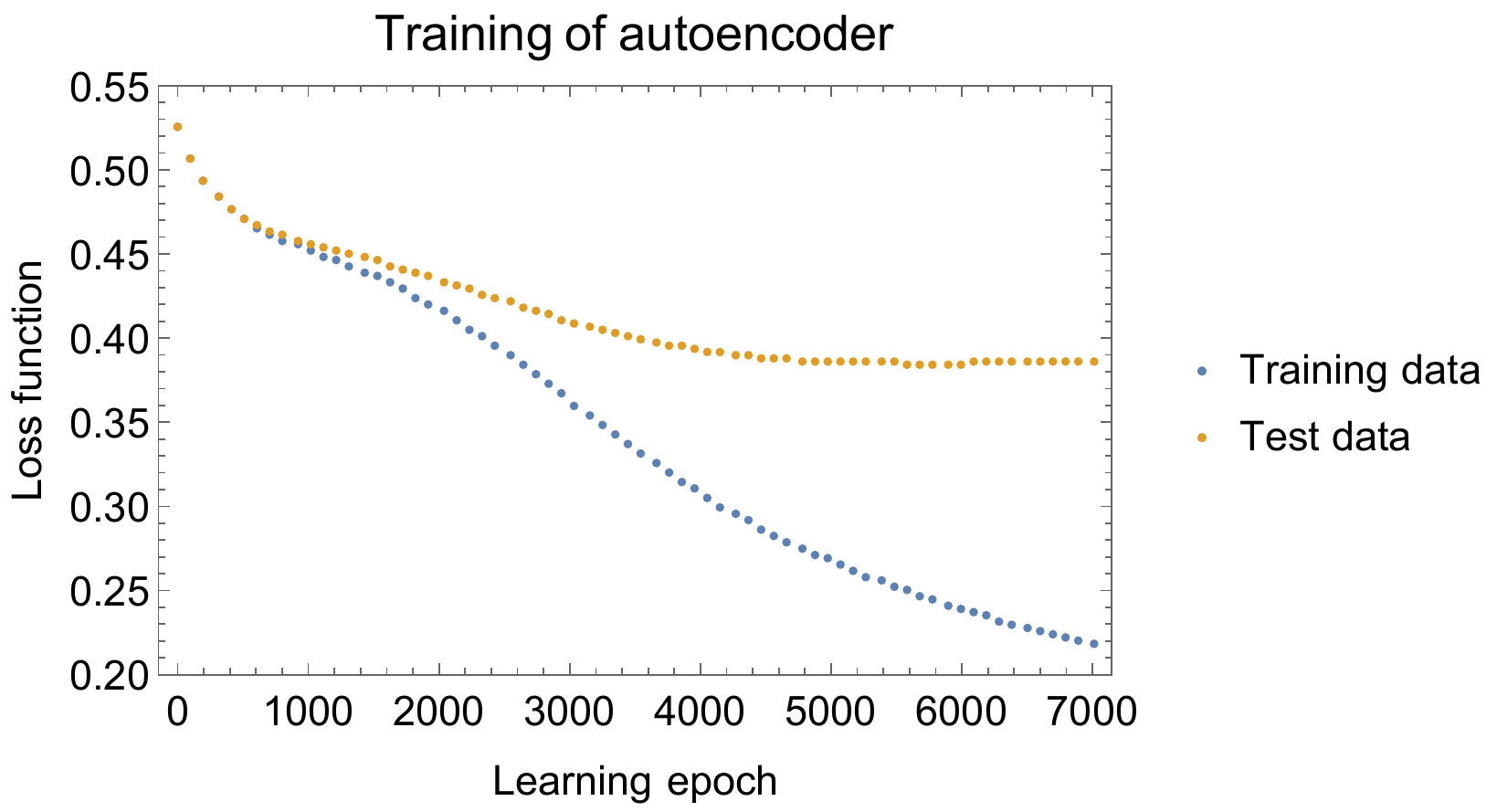}
  \caption{Loss function during training of our autoencoder.}
  \label{fig:loss_auto}
\end{figure}

As the loss function for the training, we choose the squared difference
between input data $\phi_i$ and output data $\varphi_{i,j}$,
that is,
\begin{equation}
\frac{1}{2L}\sum_{i=1}^L \left[(\re\phi_i-\varphi_{i,1})^2 + (\im\phi_i-\varphi_{i,2})^2\right].
\end{equation}
We prepared the training and test data,
both of which contain 63 winding patterns times 1000 randomly generated data. 
Then we found 
the unsupervised learning with the learning rate $10^{-7}$ and the mini-batch size 10
decreases the loss function of the test data to its minimum around 6000 epochs,
as shown in Fig.~\ref{fig:loss_auto}.

% --- figure ---%
% --- figure ---%

\bibliographystyle{apsrev4-1}
\bibliography{winding}

\end{document}